\RequirePackage{fix-cm}
\documentclass[twocolumn]{svjour3}          
\smartqed  
\usepackage{graphicx}
\usepackage[utf8]{inputenc}
\usepackage[english]{babel}
\usepackage{makeidx}
\usepackage{graphicx}
\usepackage[hidelinks]{hyperref}
\usepackage{amsmath}
\usepackage[inline]{enumitem}
\usepackage{multirow}
\usepackage{subcaption}
\usepackage{caption}
\usepackage{color}
\usepackage{booktabs}
\usepackage{siunitx}
\usepackage{algpseudocode}
\usepackage{algorithmicx}
\usepackage{array}
\usepackage{float}
\usepackage{todonotes}
\usepackage[keeplastbox]{flushend}
\usepackage[linesnumbered,ruled,vlined]{algorithm2e}
\usepackage{amsfonts}
\usepackage{threeparttable}
\usepackage{ragged2e}
\usepackage{etoolbox}
\usepackage{xcolor}
\usepackage{natbib}
\usepackage[normalem]{ulem}
\journalname{Journal of Ambient Intelligence and Humanized Computing}

\newcommand{\RVV}[1]{{\color{blue}{#1}}}
\newcommand{\RV}[1]{{\color{black}{#1}}}

\makeatletter 
\pretocmd\@bibitem{\color{black}\csname keycolor#1\endcsname}{}{\fail}
\newcommand\citecolor[1]{\@namedef{keycolor#1}{\color{black}}}
\makeatother
\citecolor{_4}
\citecolor{opixray2}
\citecolor{wang1}
\citecolor{wang2}
\citecolor{najla1}
\citecolor{najla2}

\begin{document}

\title{Unsupervised Anomaly Instance Segmentation for Baggage Threat Recognition}


\author{Taimur Hassan\textsuperscript{$\star$}         \and
        Samet Ak\c{c}ay \and
        Mohammed Bennamoun \and
        Salman Khan \and
        Naoufel Werghi
}


\institute{T. Hassan (\textsuperscript{$\star$}Corresponding Author) \at
              Center for Cyber-Physical Systems (C2PS), Department of Electrical Engineering and Computer Sciences, Khalifa University, Abu Dhabi, United Arab Emirates. \\
              \email{taimur.hassan@ku.ac.ae}           
           \and
           S. Ak\c{c}ay \at
              Department of Computer Sciences, Durham University, Durham, United Kingdom.
              \and
              M. Bennamoun \at Department of Computer Science and Software Engineering, The University of Western Australia, Perth, Australia.
              \and
              S. Khan \at Mohamed bin Zayed University of Artificial Intelligence, Abu Dhabi, United Arab Emirates.
              \and
              N. Werghi \at Center for Cyber-Physical Systems (C2PS), Department of Electrical Engineering and Computer Sciences, Khalifa University, Abu Dhabi, United Arab Emirates.
}

\date{Received: date / Accepted: date}

\maketitle

\begin{abstract}
Identifying potential threats concealed within the baggage is of prime concern for the security staff. Many researchers have developed frameworks that can automatically detect baggage threats from security X-ray scans. However, to the best of our knowledge, all of these frameworks require extensive training efforts on large-scale and well-annotated datasets, which are hard to procure in the real world, especially for the rarely seen contraband items. This paper presents a novel unsupervised anomaly instance segmentation framework that recognizes baggage threats, in X-ray scans, as anomalies without requiring any ground truth labels. Furthermore, thanks to its stylization capacity, the framework is trained only once, and at the inference stage, it detects and extracts contraband items regardless of their scanner specifications. Our one-staged approach initially learns to reconstruct normal baggage content via an encoder-decoder network utilizing a proposed stylization loss function. The model subsequently identifies the abnormal regions by analyzing the disparities within the original and the reconstructed scans. The anomalous regions are then clustered and post-processed to fit a bounding box for their localization. In addition, an optional classifier can also be appended with the proposed framework to recognize the categories of these extracted anomalies. A thorough evaluation of the proposed system on four public baggage X-ray datasets, without any re-training, demonstrates that it achieves competitive performance as compared to the conventional fully supervised methods \RV{(i.e., the mean average precision score of 0.7941 on SIXray, 0.8591 on GDXray, 0.7483 on OPIXray, and 0.5439 on COMPASS-XP dataset)} while outperforming state-of-the-art semi-supervised and unsupervised baggage threat detection frameworks \RV{by 67.37\%, 32.32\%, 47.19\%, and 45.81\% in terms of F1 score across SIXray, GDXray, OPIXray, and COMPASS-XP datasets, respectively}. 
\end{abstract}


\section{Introduction}

\noindent Recognizing contraband items concealed within baggage is a prime security concern as it endangers public safety. \RV{According to a recent report, approximately 1.5 million passengers in the United States are searched every day for weapons and other dangerous items \RVV{\citep{_4}}. Manual detection of these items is a tiring task and also subject to human errors caused due to fatigued work schedules, amount of baggage clutter, aviation traffic load, or simply because of less experience towards screening the contraband data.} To overcome this, many researchers have developed automated frameworks \RVV{\citep{gaus2019evaluating, akcay2018ganomaly, turcsany2013improving}} to screen baggage at airports, malls, and cargoes. \RV{However, the majority of these frameworks are developed using conventional RGB detectors, which have limited performance towards localizing the occluded suspicious objects due to their region based proposal generation mechanisms \RVV{\citep{hassan2019}}, and because of the inherent differences between the RGB and the X-ray imagery \RVV{\citep{akcay2018using}}. To handle this, researchers have recently proposed clutter-aware solutions possessing the capacity to recognize concealed and occluded baggage threats regardless of the scanner specifications \RVV{\citep{Hassan2020ACCV,hassan2019,hassan2020Sensors}}, acquisition noise \RVV{\citep{opixray2}}, and the imbalanced nature of the contraband data \RVV{\citep{miao2019sixray}}. In addition to this, the majority of these methods have been quantitatively evaluated to detect threatening items against different levels of occlusion on the publicly available datasets \RVV{\citep{opixray, hassan2020Sensors, Hassan2020ACCV}}. Also, the researchers have utilized 3D detectors to get rid of occlusion while screening the baggage threats from the volumetric computed tomography (CT) imagery \RVV{\citep{wang1, wang2}}. Despite these recent advancements, many state-of-the-art baggage threat detection frameworks are still based on conventional supervised learning schemes which require extensive ground truth labels to ensure robust detection performance. Some researchers have also presented semi-supervised \RVV{\citep{akcay2018ganomaly}} and unsupervised anomaly detection \RVV{\citep{akccay2019skip}} via adversarial learning. However,  such schemes require an explicit re-training process for recognizing baggage threats from different datasets. Also, they are driven by scan-level analysis for recognizing the anomalous baggage threats  and do not possess the capacity to extract and localize the threatening items within the baggage X-ray scans \RVV{\citep{akcay2018ganomaly, akccay2019skip}}.}

\noindent To address the above limitations, we present in this work a novel unsupervised anomaly instance segmentation. The proposed approach requires only one-time training, without any ground truth labels, to recognize the baggage threats. 

\section{Related Work}
\noindent Early baggage threat detection frameworks were based on conventional machine learning employing hand-engineered features \RVV{\citep{bastan2011, riffo2015automated}}. Then deep learning methods took over, proposing supervised and unsupervised strategies for recognizing the suspicious baggage content. Here, the recent \RV{approaches} have also addressed the imbalanced \RVV{\citep{miao2019sixray}} and cluttered \RVV{\citep{Hassan2020ACCV, opixray}} nature of the threatening items in X-ray scans, which are often observed in the \RV{real world} at airports, malls, and transmission cargoes. This section first gives a brief overview of some of the conventional baggage threat detection schemes, and then it sheds light on the state-of-the-art deep learning-based approaches. For an exhaustive survey, we refer the reader to the work of \RVV{\citep{ackay2020, Mery2017TMSC, Mery2020Access}}.

\subsection{Conventional Machine Learning Methods}
\noindent \RV{Earlier methods for screening baggage threats are based on handcrafted features \RVV{\citep{najla1, najla2}} and descriptors} such as SIFT \RVV{\citep{mery2016, zhang2014}}, SURF \RVV{\citep{bastan2011}}, and FAST-SURF \RVV{\citep{kundegorski2016}}, employed with Support Vector Machines (SVM) \RVV{\citep{turcsany2013improving, kundegorski2016}}, Bag of Words (BoW), K-Nearest Neighbors \RVV{\citep{riffo2015automated}} and Random Forest \RVV{\citep{jaccard2014automated}} classifiers. Many researchers have also proposed supervised segmentation \RVV{\citep{heitz2010}} and detection \RVV{\citep{bastan2015}} schemes for recognizing prohibited items via high, low and multi-view X-ray imagery \RVV{\citep{bastan2015}}. Similarly, Riffo et al. \RVV{\citep{riffo2015automated}} proposed an Adapted Implicit Shape Model (AISM) for recognizing different contraband items from the publicly available GDXray \RVV{\citep{mery2015gdxray}} dataset. In another approach, they developed structure-from-motion-based 3D feature descriptors for recognizing the threatening items \RVV{\citep{mery2016}}.

\noindent Although, traditional machine learning methods can mass-screen the baggage content using security X-ray scans. However, they are only applicable to limited experimental settings and cannot be well-generalized to multiple scanner specifications. 

\subsection{Deep Learning Methods}
\noindent \RV{Deep learning has greatly enhanced the recognition capabilities of the threat screening frameworks such that they can now identify suspicious objects within grayscale or colored baggage X-ray scans regardless of their scanner properties. Here, we categorized all the deep learning-based threat detection frameworks as supervised and unsupervised approaches.}

\subsubsection{Supervised Approaches}
\noindent Supervised approaches for recognizing baggage threats employed classification \RVV{\citep{ akccay2016transfer, jaccard2017, zhao2018, miao2019sixray}}, detection \RVV{\citep{liu2018detection, Xu2018, hassan2019, hassan2020Sensors}} and segmentation \RVV{\citep{Hassan2020ACCV, an2019, gaus2019evaluation}} strategies. \RV{Ak\c{c}ay et al. \RVV{\citep{akccay2016transfer}} introduced GoogleNet \RVV{\citep{googlenet}} (in a transfer learning mode) to detect threatening objects within baggage X-ray imagery.} Jaccard et al. \RVV{\citep{jaccard2017}} used \RV{VGG-19 \RVV{\citep{vgg16}} on log-transformed scans to detect suspicious objects.} Zhao et al. \RVV{\citep{zhao2018}} initiated the use of GANs to enhance the classification performance of the customized networks \RV{towards baggage threat detection. Apart from this, researchers have also used} two-staged \RVV{\citep{liu2018detection}} and one-staged \RVV{\citep{gaus2019evaluation}}  \RV{detectors along with} attention mechanisms \RVV{\citep{Xu2018}} \RV{to recognize and localize threatening objects. Moreover,} Gaus et al. \RVV{\citep{gaus2019evaluating}}  measured the transferability of Faster R-CNN \RVV{\citep{fasterrcnn}} Mask R-CNN \RVV{\citep{maskrcnn}} and RetinaNet \RVV{\citep{retinanet}} between various X-ray scanners \RV{to detect the contraband data.} Motivated by the class imbalance between normal and suspicious objects, Miao et al. \RVV{\citep{miao2019sixray}} presented the class-balanced hierarchical refinement (CHR) model, proposing architecture-oriented mitigation of the class imbalance problem. Other approaches proposed contour-driven object detectors such as Cascaded Structure Tensors (CST) \RVV{\citep{hassan2019}}, and  Dual-Tensor Shot Detector (DTSD) \RVV{\citep{hassan2020Sensors}}. Similarly, Wei et al. \RVV{\citep{opixray}} developed De-occlusion Attention Module (DOAM), a plug-and-play module that can be integrated with conventional object detectors to increase their capacity in screening occluded baggage threats. 
For the segmentation approaches, An et al. \RVV{\citep{an2019}} employed encoder-decoder models leveraging dual attention mechanisms, while \RVV{\citep{Hassan2020ACCV}} proposed a first-ever contour instance segmentation framework exclusively designed to extract cluttered contraband data from the security X-ray scans.

\subsubsection{Unsupervised Approaches}
\noindent Researchers have also developed semi-supervised and unsupervised methods for recognizing suspicious items. Ak\c{c}ay et al. pioneered this by developing GANomaly \RVV{\citep{akcay2018ganomaly}}, an encoder-decoder-encoder-driven adversarial framework trained on the normal security X-ray scans. After training, GANomaly \RVV{\citep{akcay2018ganomaly}} recognizes the baggage threats, as anomalies, from the abnormal test scans through its in-built discriminator. In another approach, Skip-GANomaly is proposed \RVV{\citep{akccay2019skip}} as an improved version of GANomaly utilizing encoder-decoders with skip-connections and adversarial learning to detect anomalous baggage threats with a significantly lesser amount of computational resources.

\noindent To the best of our knowledge, the majority of the existing frameworks are based on supervised learning, requiring an extensive amount of well-annotated training data to perform well at the inference stage \RVV{\citep{Hassan2020ACCV, miao2019sixray, opixray, gaus2019evaluation}}. However, procuring a large-scale and well-annotated dataset is often impractical and infeasible, especially for recognizing those items \RV{which} are rarely observed during the aviation screening. Furthermore, re-training or even fine-tuning the deployed framework to identify a new type of threat is an inefficient process and could lead to compromised performance \RVV{\citep{gaus2019evaluating, hassan2020Sensors}}. Despite recent efforts leveraging meta-transfer-learning \RVV{\citep{meta}} to alleviate the scanner differences and increase the generalizability of baggage threat detectors \RVV{\citep{hassan2020Sensors}}, these frameworks still require fine-tuning on different datasets for achieving good performance. Although researchers have utilized \RV{semi-supervised and} unsupervised adversarial learning to recognize suspicious baggage items \RVV{\citep{akcay2018ganomaly, akccay2019skip}}. These frameworks still require explicit re-training on the normal data of each dataset to identify baggage threats. Also, these methods can recognize suspicious items but are unable to localize them \RV{through bounding boxes or masks}. Hence, this paper presents the first unsupervised anomaly instance segmentation framework exclusively designed to recognize and localize illegal baggage items from the security X-ray scans to the best of our knowledge. 

\section{Contributions}
\noindent This paper presents a novel unsupervised anomaly instance segmentation framework to detect and extract baggage threats as anomalies. To the best of our knowledge, this is the first approach towards unsupervised anomaly instance segmentation, in a baggage threat detection territory, exhibiting the following distinctive features: 

\begin{itemize}

    \item The proposed framework is the first of its kind that is trained only once on the normal baggage X-ray scans. Afterward, it does not require re-training to eliminate the scanner differences to extract anomalous regions (across different datasets).\\ 
    
    \item The proposed framework is built upon a novel Gaussian-Weighted Fourier Stylization (GW-FS) scheme that drastically removes the scanner variations to achieve high generalizability towards extracting the suspicious baggage items as anomalies from the baggage X-ray scans.\\
    
    \item \RV{A thorough validation on four public X-ray datasets showcases that the proposed framework outperforms its unsupervised and semi-supervised competitors while achieving a competitive performance with the other fully supervised frameworks.}
\end{itemize}


\noindent The rest of the paper is organized as follows: Section \ref{sec:proposed} presents the proposed framework, Section \ref{sec:exp} showcases the experimental setup, Section \ref{sec:results} presents the detailed evaluation results, and Section \ref{sec:conclusion} discusses the prospects of the proposed framework and concludes the paper. 

\begin{figure*}[t]
\begin{center}
   \includegraphics[width=1\linewidth]{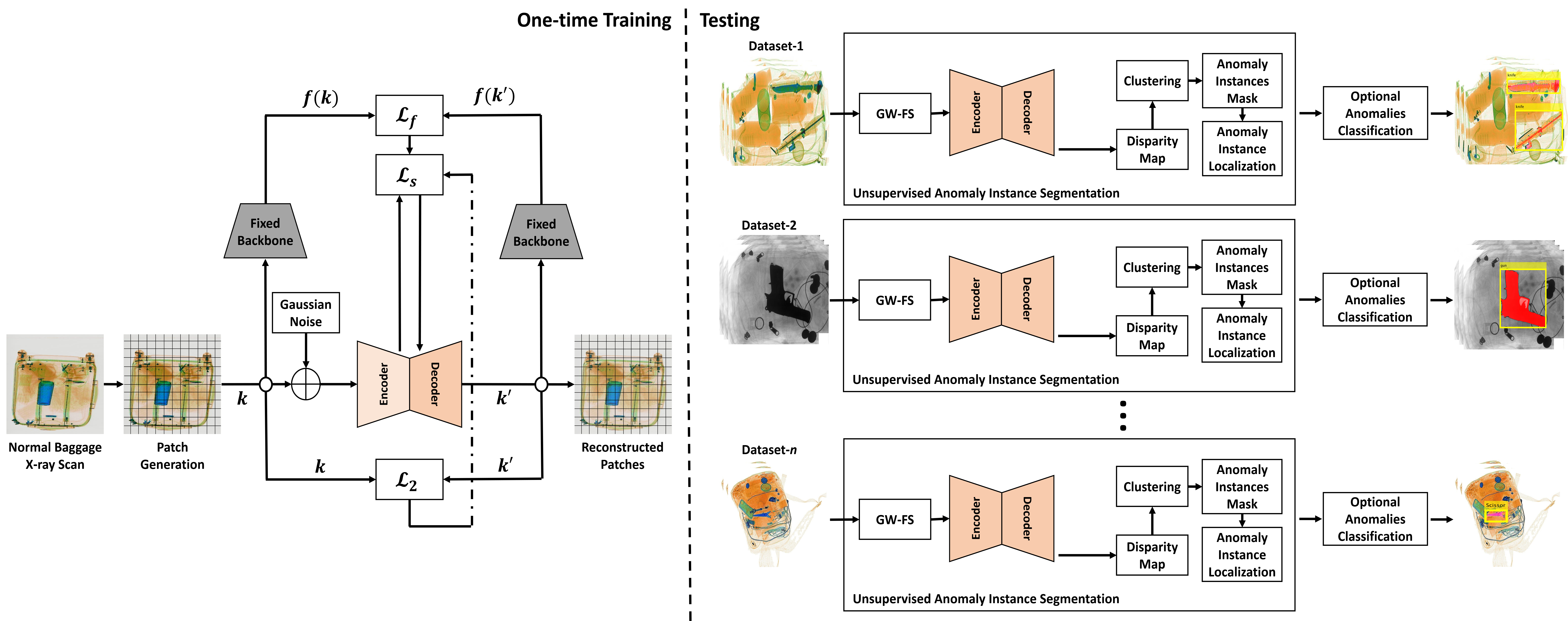}
\end{center}
   \caption{Block diagram of the proposed framework. In the one-time training stage, the X-ray scans (containing the normal baggage data) are decomposed into fixed-size non-overlapping patches, which are passed to the proposed encoder-decoder network that learns to reconstruct them. Afterward, during the inference stage, the trained model is fed with the abnormal scans. After reconstructing them, the proposed framework exploits the original and reconstructed scans' disparities to recognize anomalous regions. Furthermore, the proposed GW-FS scheme removes the scanner-specific appearance enabling the proposed framework to identify baggage threats irrespective of the scanner specifications or the dataset. }
\label{fig:fig2}
\end{figure*}


\section{Proposed Approach} \label{sec:proposed}
\noindent The block diagram of the proposed framework is shown in Figure \ref{fig:fig2}. \RV{We can see here that the encoder-decoder network is trained only once on the X-ray scans (containing the normal baggage data). During this one-time training, the network is constrained via custom stylization loss function ($\mathcal{L}_s$) to reconstruct the baggage X-ray scans accurately.} The reconstruction is performed patch-wise, and to ensure that the network maintains the \RV{spatial characteristics} of the original input scan, we perturb it, in each patch, with randomized zero-mean Gaussian noise. We empirically found that the addition of Gaussian noise within each patch puts more constraint to $\mathcal{L}_s$ in chastising the encoder-decoder network towards producing accurate reconstruction.  

\noindent Moreover, $\mathcal{L}_s$ minimizes not only the $\mathcal{L}_2$ loss but also the differences in the feature representations obtained \RV{from} the fixed backbone model. These two mechanisms enhance the encoder-decoder model's capacity for reconstructing the normal scans, leading to the generation of distinct disparity maps (for the abnormal scans) \RV{at} the inference stage. The disparity maps are then clustered together and are post-processed to detect and locate the anomalous regions. Moreover, an optional lightweight classifier can be mounted at the back of the proposed framework to recognize the localized anomalous items' categories. \RV{It should be noted here that before feeding the test scan into the proposed model, we stylize it first based upon the proposed Gaussian-Weighted Fourier Stylization (GW-FS). This stylization removes the scanner-specific appearances (even the drastic ones), enabling the encoder-decoder network to reconstruct the test scan patches accurately regardless of the scanner model.}
The detailed description of each module is presented below:

\subsection{Gaussian-Weighted Fourier Stylization}

\noindent To perform stylization, we propose a Gaussian-Weighted Fourier Stylization (GW-FS) scheme. The GW-FS is inspired from Fourier Domain Adaptation (FDA) \RVV{\citep{fda}} that computes the Fast Fourier Transform (FFT) \RVV{\citep{fft}} of the reference and target scans and copy the frequency samples within the magnitude spectrum of the reference scans to the target scan spectrum (defined by the rectangular window $\mathcal{R}$) without altering the phase spectrum \RVV{\citep{fda}}. However, in our approach, we perform stylization by mixing the low-frequency components within the candidate scan's magnitude response with the reference scan's magnitude spectrum by fitting a Gaussian window (parameterized by the variance $\sigma$). \RV{Let $x \in \mathbb{R}^{M \times N}$ be the input scan (such that $M$ denotes its height and $N$ denotes its width), and $y \in \mathbb{R}^{M \times N}$ be the reference image.} Taking FFT yields:

\begin{equation}
X_{u,v} = \frac{1}{M N} \sum_{s=0}^{M-1} \sum_{t=0}^{N-1} (-1)^{s+t} x_{s,t} e^{-j 2 \pi \left (u\frac{s}{M} + v\frac{t}{N} \right ) },
    \label{eq:eq1}
\end{equation}
\noindent where $X=\mathcal{F}\{x\}$ \RV{represents} the complex frequency spectrum of $x$\RV{, $\mathcal{F}$ denotes the FFT operator,} and the factor $(-1)^{s+t}$ shifts the image spectrum by $M/2$ and $N/2$ to center-align it. We apply the same transformation to reference scan $y$, yielding $Y=\mathcal{F}{\{y\}}$. Afterward, the magnitude spectra of $Y$, i.e., $|Y|$ is  multiplied by Gaussian window $\mathcal{G}$ to extract the low ranging spectral components for stylization: 

\begin{equation}
S_{u,v} = |Y_{u,v}| \times \mathcal{G}_{u,v} = \frac{1}{2 \pi \sigma^2} |Y_{u,v}| \times e^{-(u^2 + v^2)/2\sigma^2},
    \label{eq:eq2}
\end{equation}

\noindent $S \in \mathbb{R}^{M \times N}$ denotes the obtained stylization mask that is added to $|X|$ to produce  $|X^{'}| = |X| + S$.
Afterward, we apply the  inverse FFT ($\mathcal{F}^{-1}$) of $X^{'}$ to obtain the stylized scan  $x^{'} = \mathcal{F}^{-1} \{ X^{'} \}$, as expressed below:

\begin{equation}
x_{s,t}^{'} = \sum_{u=0}^{M-1} \sum_{v=0}^{N-1} (-1)^{u+v} X_{u,v}^{'} e^{j 2 \pi \left (s\frac{u}{M} + t\frac{v}{N} \right ) },
    \label{eq:eq3}
\end{equation}

\begin{figure}[t]
\begin{center}
   \includegraphics[width=1\linewidth]{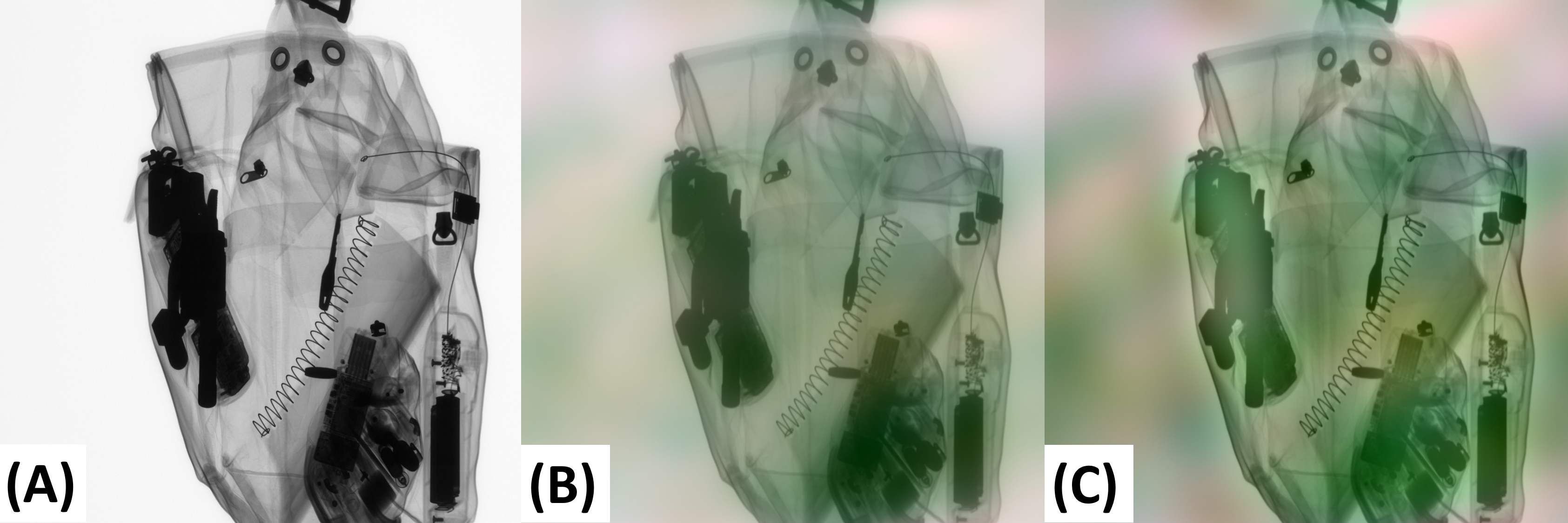}
\end{center}
   \caption{(A) Original grayscale scan, (B) stylization through GW-FS scheme with $\sigma=5$, (C) stylization through FDA \RVV{\citep{fda}} with $\beta=5$. For fairness, the value of scaling factor ($\sigma$ and $\beta$) in both schemes are chosen the same.} 
\label{fig:fig3}
\end{figure}

\noindent \RV{We can observe here that using a Gaussian window $\mathcal{G}$ instead of the rectangular window $\mathcal{R}$ (employed in FDA) results in fewer ripples within the stop-band, ensuring thus a much better stylization as evidenced in Figure \ref{fig:fig3}.}
\noindent Also, $\mathcal{R}$ (in FDA) does not blend the frequency spectrum between $|X|$ and $|Y|$. It just replaces samples within $|X|$ with that of $|Y|$ (constraint by $\mathcal{R}$), where the length of $\mathcal{R}$ is administered by the $\beta$ factor \RVV{\citep{fda}}. Therefore, the stylization through FDA \RVV{\citep{fda}} produces additional noisy artifacts (as shown in Figure \ref{fig:fig3}), and optimizing them for each training-testing combination is a  haggling job. GW-FS scheme addresses this by first weighting the frequency samples of $|Y|$ by $G$ (through Eq. \ref{eq:eq2}) before merging them with the target spectra $|X|$.

\subsection{Scans Reconstruction}
\noindent \RV{After stylizing the test scan, it is passed to the asymmetric one-time trained encoder-decoder model, which generates the reconstructed images ($x^{''}$) patch-wise. Afterward,} $x^{''}$ is utilized in developing the disparity maps for anomaly instance segmentation. The proposed encoder-decoder \RV{network} is a lightweight model containing one input layer, seven convolution layers with ReLU activations, three max-pooling, and three up-sampling layers. Furthermore, it has around 4,923 trainable parameters. For more architectural details about the proposed encoder-decoder architecture, we refer the reader to the source code repository\footnote{\label{note1} The source code of the proposed framework, along with the complete documentation is available at \url{https://github.com/taimurhassan/anomaly}.}.

\noindent Moreover, to train the proposed encoder-decoder \RV{network}, we used the proposed stylization loss function ($\mathcal{L}_s$) to constrain it, at the training time, to recognize shape, context, and edge feature appearances from the latent space vectors. The $\mathcal{L}_s$ loss function is further discussed in the subsequent section below. 

\subsubsection{The \textbf{$\mathcal{L}_s$} Loss Function}
\noindent To train the proposed encoder-decoder model, we propose a novel stylization loss ($\mathcal{L}_s$) \RV{which is}  is a linear combination of feature reconstruction loss function ($\mathcal{L}_f$) \RVV{\citep{perceptual}} and the conventional $\mathcal{L}_2$ loss function.
\begin{equation}
\mathcal{L}_s = \alpha_1 \mathcal{L}_f + \alpha_2 \mathcal{L}_2,
    \label{eq:eq4}
\end{equation}
\noindent  where $\alpha_{1,2}$ represent the loss weights. $\mathcal{L}_f$ is generated from the feature representations obtained from the frozen pre-trained backbones, and $\mathcal{L}_2$ is obtained using the pixel-wise difference between the training scan ($k$) and the reconstructed version ($k^{'}$), as expressed in Eq. \ref{eq:eq5} and \ref{eq:eq6}:
\begin{equation}
\mathcal{L}_f = \frac{1}{b_s} \sum\limits_{i=0}^{b_s} |f(k_{i}) - f(k^{'}_{i})|,
    \label{eq:eq5}
\end{equation}
\noindent 
\begin{equation}
\mathcal{L}_2 = \frac{1}{b_s} \sum\limits_{i=0}^{b_s} |k_{i} - k^{'}_{i}|^2,
    \label{eq:eq6}
\end{equation}

\noindent where $b_s$ \RV{is} the batch size, and \RV{$f(.)$ denotes the feature representations obtained from the frozen backbone model.} The loss weights $\alpha_1$ and $\alpha_2$ are empirically determined to be 0.7 and 0.3, respectively.

\subsubsection{Unsupervised Anomaly Instance Segmentation}
\begin{figure}[t]
\begin{center}
   \includegraphics[width=1\linewidth]{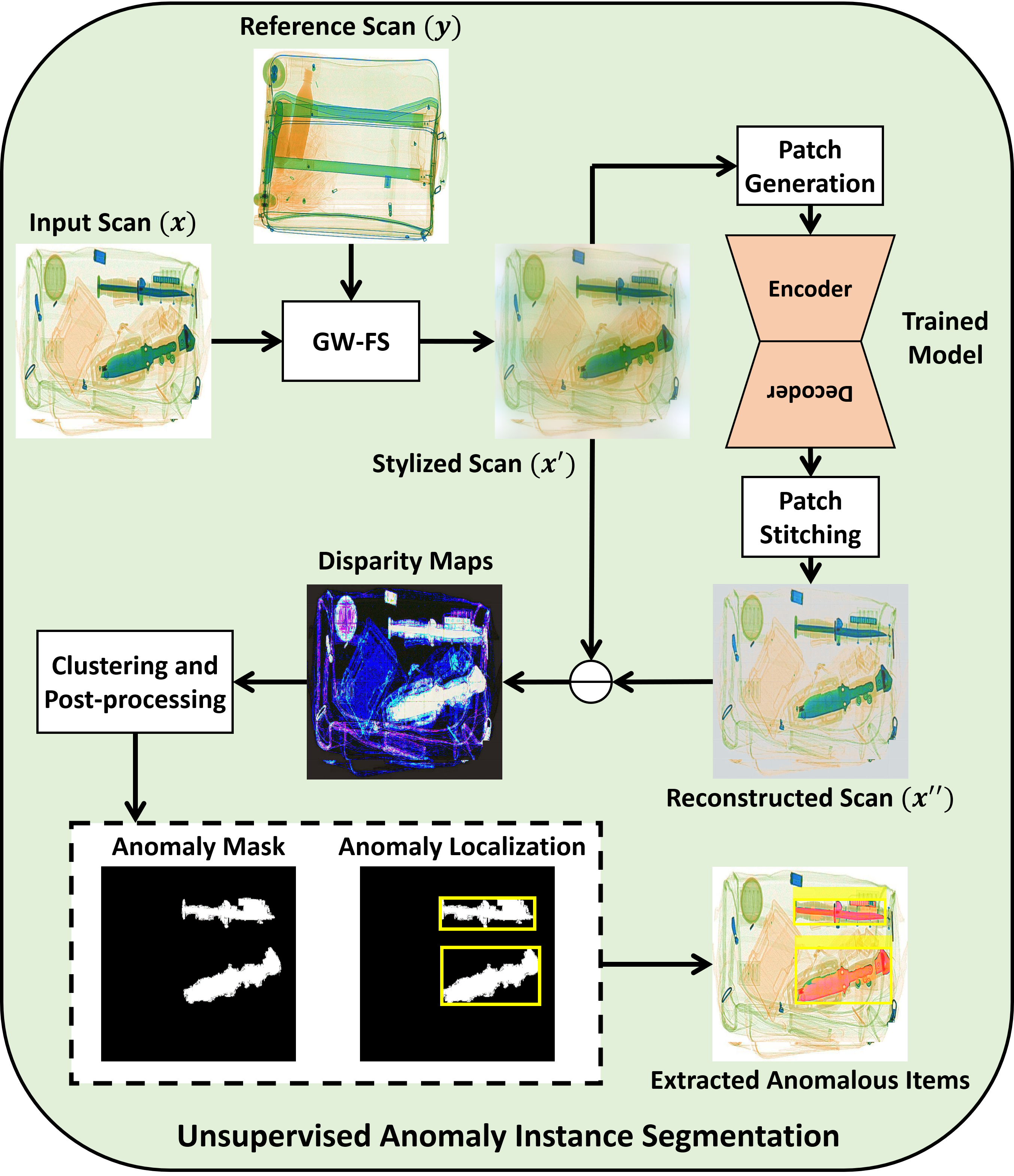}
\end{center}
   \caption{Unsupervised anomaly instance segmentation framework. Here, we train an encoder-decoder model only once to reconstruct normal baggage X-ray scans. During the inference stage, the trained model recognizes the anomalous regions by exploiting the actual and reconstructed scans' disparities. To eliminate the scanner variations, we propose a GW-FS scheme that mixes the frequency representations within the reference scan and the input scans. }
\label{fig:fig1}
\end{figure}

The proposed unsupervised anomaly instance segmentation \RV{scheme} is shown in Figure \ref{fig:fig1}.  At the inference stage, after stylizing the input scan, it is patch-wise reconstructed through the trained encoder-decoder model. Then, the reconstructed scan ($x^{''}$) is subtracted from the stylized scan ($x^{'}$) to produce the disparity map. \RV{The} disparity map is utilized \RV{in extracting} the suspicious items' instances by clustering the color distribution between anomalous and normal baggage content. The detailed description of disparity maps and the color distribution-based clustering scheme is presented in the subsequent sections.

\begin{figure}[t]
\begin{center}
    (A)\includegraphics[width=0.43\linewidth]{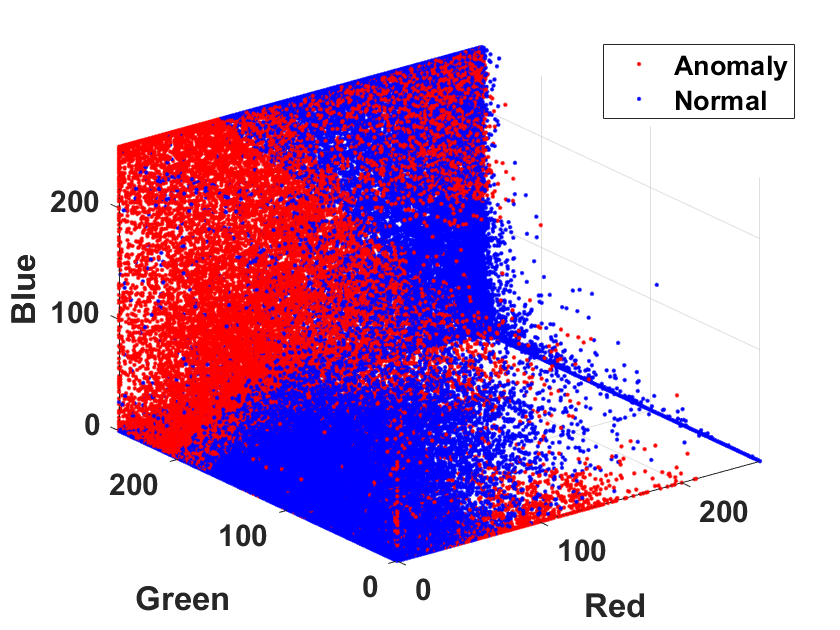}
    (B)\includegraphics[width=0.43\linewidth]{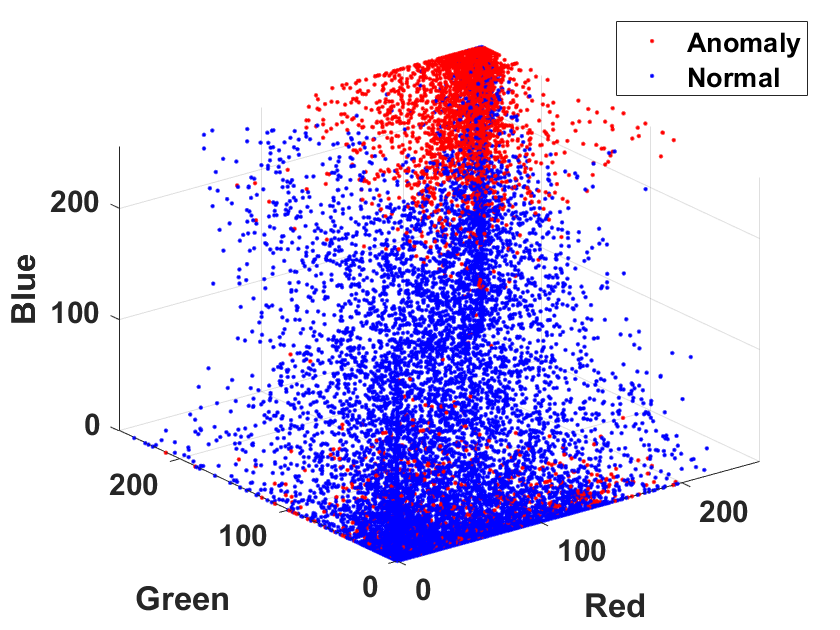}
    (C)\includegraphics[width=0.43\linewidth]{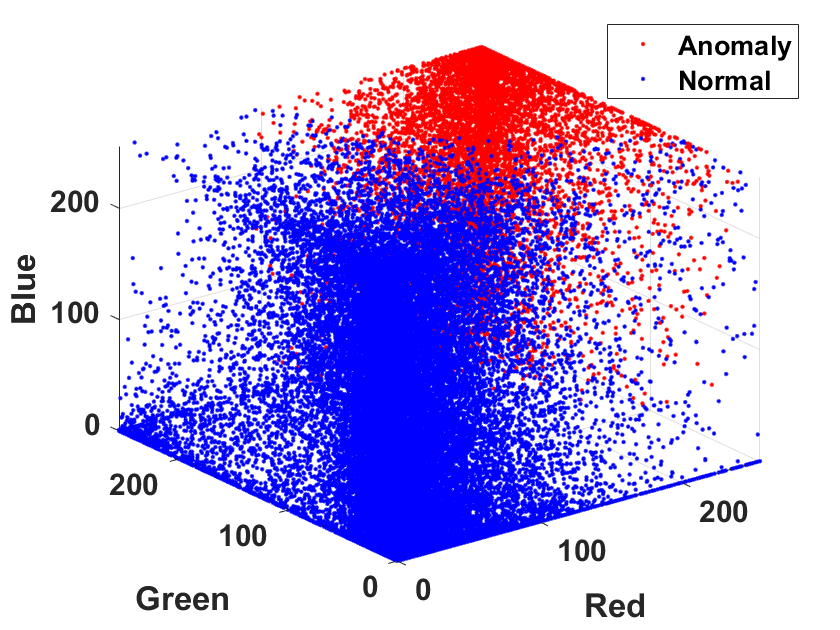}
    (D)\includegraphics[width=0.43\linewidth]{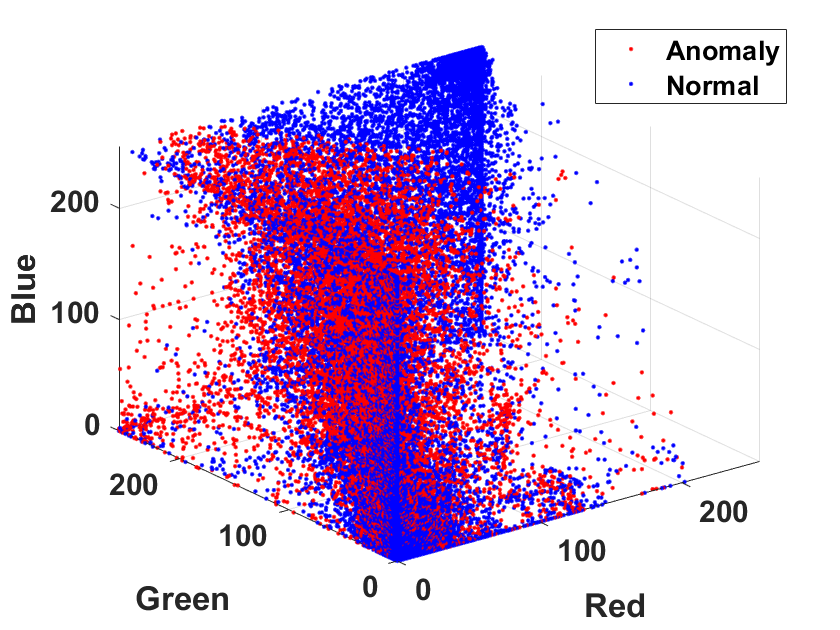}
\end{center}
   \caption{Color distributions between anomalous and non-anomalous regions in (A) SIXray \RVV{\citep{miao2019sixray}} dataset, (B) GDXray \RVV{\citep{mery2015gdxray}} dataset, (C) OPIXray \RVV{\citep{opixray}} dataset, and (D) COMPASS-XP \RVV{\citep{compass}} dataset.}
\label{fig:distribution}
\end{figure}

\noindent \textbf{Disparity Maps:}
\noindent The disparity maps reveal the deviation of the anomalous items w.r.t the normal baggage content by subtracting $x^{''}$ from $x^{'}$. It should be noted here that the meaningful interpretation from these disparity maps is subject to how accurately the encoder-decoder model reconstructs the normal areas within the abnormal scans. For example, in Figure \ref{fig:fig1}, we can see how effectively the encoder-decoder has reconstructed the abnormal scan (containing \textit{knives}). However, there are still some noticeable intensity variations between $x^{'}$ and $x^{''}$, which results in the blue, pink, and cyan color noisy regions within the disparity maps. Having a three-channeled representation here allows better discrimination between anomalous regions (corresponding to suspicious items) and the rest of the baggage content as compared to the single-channeled representations. This aspect is further evidenced in Figure \ref{fig:distribution}. Here, for each dataset, red points indicate anomalies, whereas blue color showcases normal pixels. We can observe that the distributions of anomalous and normal baggage content are well-separated. Therefore using an adequate clustering scheme, the anomalous region representing the suspicious items can be extracted. 

\noindent \textbf{Color Clustering:}
\noindent In order to \RV{extract suspicious (anomalous) items' instances}, the disparity maps are clustered through K-means Clustering (parameterized by the number of clusters $\mathcal{C}$). Here, $\mathcal{C}$ for each dataset varies depending upon normal and anomalous items contained within the respective scans. 
Through empirical analysis, we determined the optimal choice of $\mathcal{C}$ for SIXray \RVV{\citep{miao2019sixray}} and OPIXray \RVV{\citep{miao2019sixray}} dataset is four (i.e., $\mathcal{C}=4$). Similarly, for GDXray \RVV{\citep{mery2015gdxray}} and COMPASS-XP \RVV{\citep{compass}} dataset, $\mathcal{C} = 3$. 
\noindent Moreover, the noisy clusters (obtained after K-means clustering) are automatically removed through morphological post-processing. Afterward, each isolated instance of the anomalous region is identified through the connected-component analysis, and then each item instance is localized by fitting the bounding box generated using the minimum and maximum of their masks in both image dimensions \RV{(see Figure \ref{fig:fig1})}. 

\begin{figure*}[t]
\begin{center}
   \includegraphics[width=1\linewidth]{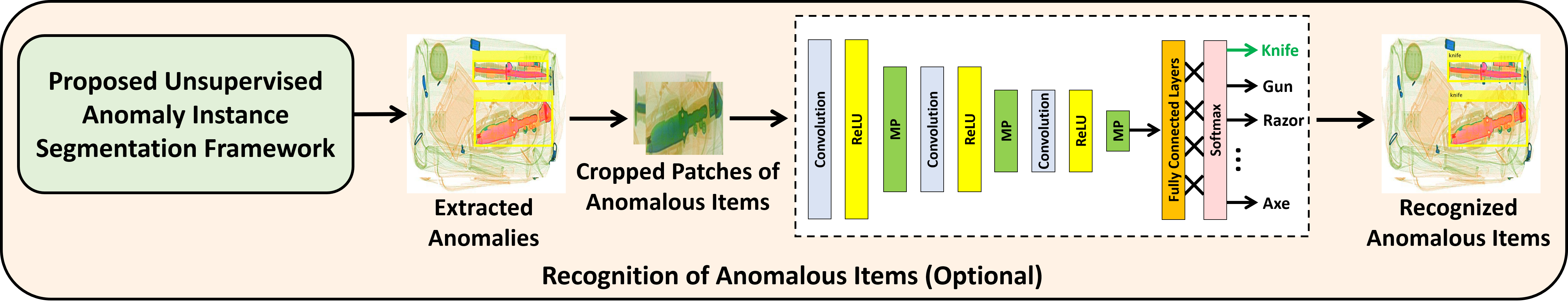}
\end{center}
   \caption{\RV{Recognition of anomalous items' categories using the proposed classification model. The model is trained only once on the suspicious items patches obtained from all four datasets.}}
\label{fig:recognition}
\end{figure*}

\subsection{Recognition of Anomalous Items}
\noindent After extracting the anomalous items we identify  their categories (such as \textit{gun}, \textit{knife}, \textit{razor}, \textit{shuriken}, \textit{wrench}, \textit{pliers}, \textit{scissor}, \textit{hammer}, and \textit{axe} etc.) using a  proposed lightweight classification model (see  Figure \ref{fig:recognition}). We want to highlight here that the recognition of anomalies items is just an optional module. It neither relates to our actual unsupervised anomaly instance segmentation approach nor it is mandatory in the proposed framework. 

\noindent The patches of the anomalous items are cropped from $x^{'}$ using their bounding boxes, and the classifier is trained only once to recognize their categories. The data used to train this model is based on patches (representing each suspicious item category), and it is taken from all four datasets.  \RV{Architecturally, the classification model contains three convolutions, three ReLUs, three max-pooling, one fully connected, and one softmax layer as depicted in Figure \ref{fig:recognition}.} The total number of parameters within the proposed model is 3.2M. Note that any pre-trained model can be used here \RV{for classifying the suspicious object categories}. However, our \RV{proposed model} exhibits two advantages 1) it is lightweight compared to the popular pre-trained models, and 2) it achieves a good trade-off between accuracy and the number of hyper-parameters (as evidenced from Table \ref{tab:tab4}). The classification model is optimized via the cross-entropy loss function ($\mathcal{L}_{c}$), as expressed below:

\begin{equation}
\mathcal{L}_{c} = -\frac{1}{b_s}\sum\limits_{i=0}^{b_s-1}\sum\limits_{j=0}^{n_c-1} t_{i,j}\log(p(l_{i,j}))
\end{equation}

\noindent where $n_c$ denotes the number of suspicious item categories, $t_{i,j}$ denotes the $i^{th}$ training example for the $j^{th}$ class, and $p_{i,j}$ denotes the softmax probability for the $i^{th}$ training sample belonging to $j^{th}$ class. Moreover, the training details of this optional classification model is presented in Section \ref{sec:training}.

\section{Experimental Setup} \label{sec:exp}
\noindent This section presents the details about the datasets, the training protocol, as well as the evaluation metrics:

\subsection{Datasets} 
The proposed framework is thoroughly evaluated on all the four public X-ray datasets used in baggage threat recognition benchmarking. The detailed description of these datasets is as follows:

\subsubsection{SIXray}
SIXray \RVV{\citep{miao2019sixray}} is the largest and most challenging baggage X-ray dataset to date. It contains 1,050,302 negative X-ray scans containing only the normal baggage content, and 8,929 positive scans having one or more suspicious items in it such as \textit{guns}, \textit{knives}, \textit{wrenches}, \textit{pliers}, \textit{scissors}, and \textit{hammers}. Furthermore, the dataset also contains detailed box-level annotations to train and evaluate the baggage threat detection frameworks. Moreover, SIXray \RVV{\citep{miao2019sixray}} is primarily designed to test the capacity of autonomous frameworks to screen contraband items in a highly imbalanced scenario. 

\subsubsection{GDXray}
The GDXray \RVV{\citep{mery2015gdxray}} dataset contains 19,407 high resolution grayscale X-ray scans divided into five groups, namely, \textit{welds}, \textit{casting}, \textit{baggage}, \textit{nature}, and \textit{settings}. The only relevant category for the proposed study is the baggage category that contains 8,150 X-ray scans along with detailed markings. Moreover, the scans within the GDXray \RVV{\citep{mery2015gdxray}} contains suspicious items such as \textit{razors}, \textit{handguns}, \textit{knives}, and \textit{shuriken} \RVV{\citep{mery2015gdxray}}.  

\subsubsection{OPIXray}
The OPIXray \RVV{\citep{opixray}} is the most recent publicly released baggage X-ray dataset. It contains a total of 8,885 colored X-ray scans containing suspicious items such as \textit{folding knives}, \textit{straight knives}, \textit{utility knives}, \textit{multi-tool knives}, and \textit{scissors} \RVV{\citep{opixray}}. Furthermore, OPIXray \RVV{\citep{opixray}} also contain the detailed box-level annotations for these items which can be used in order to evaluate the autonomous baggage threat detection frameworks.

\subsubsection{COMPASS-XP}
Different from the previous datasets, COMPASS-XP \RVV{\citep{compass}} is mainly designed to assess classification (rather than detection) frameworks, i.e., it contains scan-level markings to recognize baggage threats without ground truths masks for the localization.
However, the novel aspect of the COMPASS-XP dataset \RVV{\citep{compass}} is that it contains different scanner images for each case. For example, for a single scene in which a baggage contains a \textit{gun}, the COMPASS-XP gives six different scanner representations (i.e., the high-energy X-ray, low-energy X-ray, high density, colored X-ray, grayscale X-ray, and the normal RGB scan). So, in total, the complete dataset contains 11,568 = 6 x 1928 X-ray scans \RVV{\citep{compass}}.

\subsection{Training Details} \label{sec:training}
\noindent The proposed framework has been implemented using Python 3.7.8 with TensorFlow 2.3.0 and the MATLAB R2020a on a machine with Intel Core i9-10940@3.3 GHz processor and 132 GB RAM with a single NVIDIA Quadro RTX 6600, cuDNN v7.5, and a CUDA Toolkit v11.0.221. The training was conducted for 200 epochs using 80\% of the normal baggage X-ray scans \RV{(i.e., 840,241 normal scans)} from the SIXray dataset. The choice of the SIXray dataset for one-time training is driven from an extensive ablation study (presented in Section \ref{sec:ablationData}). \RV{Moreover, the total number of test scans from all four datasets which we used for evaluating the proposed framework are 238,664 (8,150 scans are taken from GDXray, 210,061 scans are taken from SIXray, 8,885 scans are taken from OPIXray, and 11,568 scans are taken from COMPASS-XP dataset). Apart from this, we used 8,929 scans from the SIXray dataset for validation purposes.} 
\noindent Furthermore, the optimizer used during the training was ADAM \RVV{\citep{adam}} with default learning and decay rates. For recognition of anomalous items, we trained the modular classification model for 100 epochs with ADAM \RVV{\citep{adam}} having an initial learning rate of 0.0001. The total training patches we used to train this classifier are around 5,000, obtained from all four X-ray datasets for each suspicious item category. The source code of the proposed framework is also released publicly for the research community\textsuperscript{\ref{note1}}.

\subsection{Evaluation Metrics} \label{sec:metrics}
\noindent We used standard object detection, instance segmentation, and classification metrics such as mAP, MSE, accuracy, recall, precision, F1 to test the proposed framework's performance and compare it with the state-of-the-art solutions. 

\section{Results} \label{sec:results}

\subsection{Ablation Study}
\noindent The ablation study for the proposed framework includes 1) The choice of $\sigma$ for GW-FS stylization; 2) The optimal backbone network for computing $\mathcal{L}_f$; 3) The optimal classification backbone model, and 4) the choice of training dataset which is used for performing one-time training.

\noindent \subsubsection{Choice of $\sigma$:}
\noindent The $\sigma$ is a hyper-parameter within the GW-FS scheme to determine the Gaussian window's cut-off frequency. Increasing the value of $\sigma$ increases the pass-band region and allows more frequencies to pass through, whereas decreasing the value of $\sigma$ only allows the lowest frequencies (among all) to remain while the rest are clipped. Table \ref{tab:tab1} reports the performance of GW-FS for reconstructing three-channeled scans in terms of MSE scores. Here, we can observe that for $\sigma=5$, the proposed framework achieves the minimum reconstruction error for all datasets. However, when $\sigma$ increases, the reconstruction performance starts to deteriorate because higher frequencies are being allowed to pass through the window (defined by $\sigma$), which generates more noisy edges. 

\begin{table}[b]
    \centering
    \caption{Effects of varying $\sigma$ on scan reconstruction. Bold indicates the best MSE scores.} 
    \begin{tabular}{ccccc}
    \hline
         $\sigma$ & SIXray & GDXray  & OPIXray & COMPASS-XP \\\hline 
          5 & \textbf{72.92} & \textbf{16.29} & \textbf{21.12} & \textbf{21.94} \\ 
          10 & 89.16 & 24.83 & 41.96 & 43.62 \\
          25 & 159.53 & 73.92 & 119.65 & 136.42 \\
          50 & 235.98 & 134.76 & 216.94 & 218.16 \\\hline
    \end{tabular}
    \label{tab:tab1}
\end{table}

\noindent \subsubsection{Backbone Network for Computing $\mathcal{L}_f$:}
\noindent This ablation study reports the backbone's choice for computing the feature reconstruction loss function ($\mathcal{L}_f$). Here, we compared the performance of pre-trained VGG-16 \RVV{\citep{vgg16}} and ResNet-50 \RVV{\citep{he2016deep}} that produces $\mathcal{L}_f$ to penalize the encoder-decoder model towards reconstructing the abnormal scans robustly. It should be noted here that these pre-trained models were fixed, i.e., the weights of these networks were not trained explicitly for minimizing $\mathcal{L}_f$. The results for this experiment are reported in Table \ref{tab:tab2}. We can see here that $\mathcal{L}_f$ with VGG-16 \RVV{\citep{vgg16}} achieves 9.03\% better performance on SIXray \RVV{\citep{miao2019sixray}} dataset. Similarly, on GDXray \RVV{\citep{mery2015gdxray}}, OPIXray \RVV{\citep{opixray}}, and COMPASS-XP \RVV{\citep{compass}} dataset, VGG-16 \RVV{\citep{vgg16}} driven $\mathcal{L}_f$ achieved 20.69\%, 24.08\%, and 31.67\% better reconstruction performance, respectively. However, if we increase the training epochs for ResNet-50 \RVV{\citep{he2016deep}}, we can achieve similar performance. But since VGG-16 \RVV{\citep{vgg16}} produced better results with lesser training, we opted for it in the proposed framework to compute $\mathcal{L}_f$ during one-time training. 

\begin{table}[htb]
    \centering
    \caption{Performance of different pre-trained models for computing $\mathcal{L}_f$ during one-time training. Bold indicates the best MSE scores. Moreover, the abbreviation `COMP' represents the COMPASS-XP dataset \RVV{\citep{compass}}.} 
    \begin{tabular}{ccccc}
    \hline
          Model & SIXray & GDXray & OPIXray & COMP \\\hline
          VGG-16 & \textbf{72.92} & \textbf{16.29} & \textbf{21.12} & \textbf{21.94} \\
          ResNet-50 & 80.16 & 20.54 & 27.82 & 32.11 \\\hline
    \end{tabular}
    \label{tab:tab2}
\end{table}

\noindent \subsubsection{The Optimal Classification Backbone:}
\noindent Recognition of anomalous item (after unsupervised anomaly instance segmentation) is an optional step required to identify the type of anomaly contained within the localized anomalous region. To perform this, we exploited different pre-trained models (fine-tuned on suspicious items patches). We also compared the performance of a proposed classification model with these pre-trained models to see how well it recognizes the suspicious items (contained within the patches). The comparison is reported in Table \ref{tab:tab4}. Here, we can observe that with few training examples (i.e., the suspicious items patches) from all the datasets, the proposed model achieves competitive classification performance compared to other pre-trained networks. Furthermore, considering that it has 54.48\% fewer parameters than best performing DenseNet-121 \RVV{\citep{densenet}} model. We believe that it provides a good trade-off between performance and computational requirements, especially compared to the MobileNetv2 \RVV{\citep{mobilenet}}. 

\begin{table}[b]
    \centering
    \caption{Comparison of classification performance for recognizing anomalous items patches. The good trade-off between classification performance and the computational requirements is highlighted in bold. Moreover, the abbreviations are ACC: Accuracy, TPR: True Positive Rate, PPV: Positive Predicted Value, F1: F1 Score, NP: Number of Parameters, PB: Proposed Backbone, V-16: VGG-16 \RVV{\citep{vgg16}}, R-50: ResNet-50 \RVV{\citep{he2016deep}}, R-101: ResNet-101 \RVV{\citep{he2016deep}}, R-152: ResNet-152 \RVV{\citep{he2016deep}}, D-121: DenseNet-121 \RVV{\citep{densenet}}, MNv2: MobileNetv2 \RVV{\citep{mobilenet}}.} 
    \begin{tabular}{cccccc}
    \hline
          Model & ACC & TPR & PPV & F1 & NP \\\hline
          PB & \textbf{0.9669} & \textbf{0.8683} & \textbf{0.5083} & \textbf{0.6412} & \textbf{3.2M}\\
          V-16 & 0.9630 & 0.8538 & 0.4759 & 0.6111 & 14.7M\\
          R-50 & 0.9740 & 0.9172 & 0.5745 & 0.7064 & 23.5M \\
          R-101 & 0.9786 & 0.9341 & 0.6242 & 0.7483 & 42.6M \\
          R-152 & 0.9877 & 0.9446 & 0.7568 & 0.8403 & 58.3M \\
          D-121 & 0.9754 & 0.9053 & 0.5909 & 0.7150 & 7.03M \\
          MNv2 & 0.9527 & 0.8247 & 0.4049 & 0.5431 & 2.2M \\\hline
    \end{tabular}
    \label{tab:tab4}
\end{table}

\subsubsection{Choice of One-Time Training Dataset} \label{sec:ablationData}
\noindent The encoder-decoder model within the proposed framework is trained only once, and in this one-time training, it learns to reconstruct the normal baggage content at run-time robustly. As the model does not learn to recognize suspicious objects (during training), it faces a hard time reconstructing them at the inference stage, and this is highlighted within the disparity maps. 

\noindent In this experiment, we determine the optimal choice of the dataset for training the encoder-decoder model. The reconstruction performance of the proposed model (in terms of MSE scores) is reported in Table \ref{tab:trainingData}. Here, we can see that using SIXray \RVV{\citep{miao2019sixray}} dataset for training; the proposed model produces the best reconstruction performance across all four datasets at the inference stage. This is because SIXray \RVV{\citep{miao2019sixray}}, to the best of our knowledge, contains the maximum amount of scans depicting diverse ranging normal baggage content, allowing the encoder-decoder model to learn the unique feature representations within the baggage X-ray scans robustly. 
Training on GDXray \RVV{\citep{mery2015gdxray}} dataset does not produce a very good performance for two reasons: 1) It contains significantly lesser normal baggage scans (around 1,130 of them) for training purposes. 2) GDXray is a grayscale dataset, thus styling the colored baggage X-ray scans as grayscale would create more ambiguities between normal and abnormal baggage content, resulting in the noisier disparity maps. We did not use OPIXray \RVV{\citep{opixray}} dataset for training purposes because it does not contain any normal baggage X-ray scans \RVV{\citep{opixray}}. Also, COMPASS-XP \RVV{\citep{opixray}} dataset does not have complex X-ray scans (like other datasets), i.e., it contains single-item scans, and the model trained on these scans does not get much exposure towards learning diversified feature representations contained within other datasets.       

\begin{table}[t]
    \centering
    \caption{Reconstruction performance of the proposed encoder-decoder model in terms of MSE scores when trained on different datasets. The abbreviation `COMP' represents the COMPASS-XP dataset \RVV{\citep{compass}}.} 
    \begin{tabular}{cccccc}
    \hline
          Training & SIXray & GDXray & OPIXray & COMP\\
          Dataset &&&&\\\hline 
          SIXray & 72.928 & 16.291 & 21.125 & 21.941 \\ 
          GDXray & 96.532 & 13.639 & 45.923 & 51.649 \\ 
          COMP & 81.692 & 21.638 & 30.113 & 10.582 \\ 
          \hline
    \end{tabular}
    \label{tab:trainingData}
\end{table}

\subsection{Comparison with Supervised Frameworks}
\noindent In this series of experiments, we compared the proposed framework's detection and recognition performance with the state-of-the-art fully supervised baggage threat detection frameworks. Here, our approach is semi-supervised (since we mounted the optional classification module with the proposed framework for recognizing the suspicious items' categories). The comparison is reported in Table \ref{tab:tab5} where we can see that although the proposed framework lags from some state-of-the-art methods, its performance is still quite appreciable, especially considering the fact that it is a semi-supervised approach, unlike its conventionally trained, fully supervised competitors. Also, it achieves the good detection performance (i.e., it only lags from the best performing framework by 17.23\%, 11.17\%, 0.65\%, and 6.89\% on SIXray, GDXray, OPIXray, and COMPASS-XP dataset, respectively, in terms of mAP scores). \RV{Furthermore, compared to recently introduced GBAD \RVV{\citep{gbad}}, DTSD \RVV{\citep{hassan2020Sensors}}, and DOAM-O \RVV{\citep{opixray2}} approaches, the proposed framework, on SIXray and OPIXray datasets, achieve 5.76\%, 18.68\%, and 0.347\% improvements, respectively which is quite significant.}

\begin{table}[t]
    \centering
    \caption{Comparison of proposed approach (semi-supervised version) with existing fully supervised baggage threat detection frameworks in terms of mAP. Bold indicates the best score while the second-best scores are underlined. '-' indicates that the metric is not computed. Moreover, the abbreviations are COMP: COMPASS-XP \RVV{\citep{compass}}, TST: Trainable Structure Tensors \RVV{\citep{Hassan2020ACCV}}, CST: Cascaded Structure Tensors \RVV{\citep{hassan2019}}, DTSD: Dual-Tensor Shot Detector \RVV{\citep{hassan2020Sensors}}, DOAM: De-occlusion Attention Module \RVV{\citep{opixray}} with Single-Shot Detector \RVV{\citep{Liu2016SSD}}, GBAD: GAN Based Anomaly Detection \RVV{\citep{gbad}} with ResNet-101 \RVV{\citep{he2016deep}}\RV{, and DOAM-O: Oversampling De-occlusion Attention Module \RVV{\citep{opixray2}} with Single-Shot Detector \RVV{\citep{Liu2016SSD}}.}} 
    \begin{tabular}{ccccc}
    \hline
          Model & SIXray & GDXray & OPIXray & COMP \\\hline
          Proposed & 0.7941 & 0.8591 & \underline{0.7483} & \underline{0.5439} \\
          TST & \underline{0.9516} & \textbf{0.9672} & \textbf{0.7532} & \textbf{0.5842} \\
          CST & \textbf{0.9595} & \underline{0.9343} & - & - \\
          DTSD & 0.6457 & 0.9162 & - & - \\
          DOAM & - & - & 0.7401 & - \\
          GBAD & 0.7483 & - & - & - \\
          \RV{DOAM-O} & \RV{-} & \RV{-} & \RV{0.7457} & \RV{-} \\
          \hline
    \end{tabular}
    \label{tab:tab5}
\end{table}

\subsection{Comparison with Unsupervised Frameworks}
\noindent We also compared the performance of the proposed unsupervised framework with state-of-the-art methods such as GANomaly \RVV{\citep{akcay2018ganomaly}}, and Skip-GANomaly \RVV{\citep{akcay2018ganomaly}}. Here, the experimental protocol is to classify the abnormal vs. normal baggage X-ray scans (except for the OPIXray), where abnormal scans are those scans that contain one or more anomalous regions, and the normal scans only have the normal baggage content. For the OPIXray dataset, we followed the strategy of classifying the scans as having \textit{folding knives}, \textit{utility knives}, \textit{straight knives}, \textit{multi-tool knives}, and \textit{scissors}, since this dataset does not contain any normal baggage X-ray scans \RVV{\citep{opixray}}. Moreover, for fairness, all the frameworks were trained on a single SIXray dataset as per the training protocol defined in Section \ref{sec:training}. Similarly, they are also applied to the other datasets in a zero-shot manner (without any re-training or fine-tuning). The comparison is reported in Table \ref{tab:tab6} in terms of F1 scores where we can see that the proposed framework has 67.37\% lead on SIXray dataset, 32.32\% lead on GDXray dataset, and 45.81\% lead on COMPASS-XP dataset. Furthermore, on the OPIXray dataset, the proposed framework is leading the second-best Skip-GANomaly \RVV{\citep{akccay2019skip}} by 47.19\%.

\begin{table}[t]
    \centering
    \caption{Comparison of proposed framework with state-of-the-art unsupervised baggage threat detection frameworks in terms of F1 score. Bold indicates the best scores while the second-best are underlined. Moreover, the abbreviations are: PF: Proposed Framework, GA: GANomaly \RVV{\citep{akcay2018ganomaly}}, SG: Skip-GANomaly \RVV{\citep{akccay2019skip}}, and COMP: COMPASS-XP \RVV{\citep{compass}}.} 
    \begin{tabular}{ccccc}
    \hline
          Model & SIXray & GDXray  & COMP & OPIXray\\\hline
          PF & \textbf{0.4831} & \textbf{0.7839} & \textbf{0.4119} & \textbf{0.6560}\\
          GA & 0.1227 & 0.4994 & 0.2405 & 0.3074\\
          SG & \underline{0.1576} & \underline{0.5305} & \underline{0.2232} & \underline{0.3464}\\\hline
    \end{tabular}
    \label{tab:tab6}
\end{table}

\subsection{Qualitative Evaluations}
\begin{figure*}[t]
\begin{center}
   \includegraphics[width=0.99\linewidth]{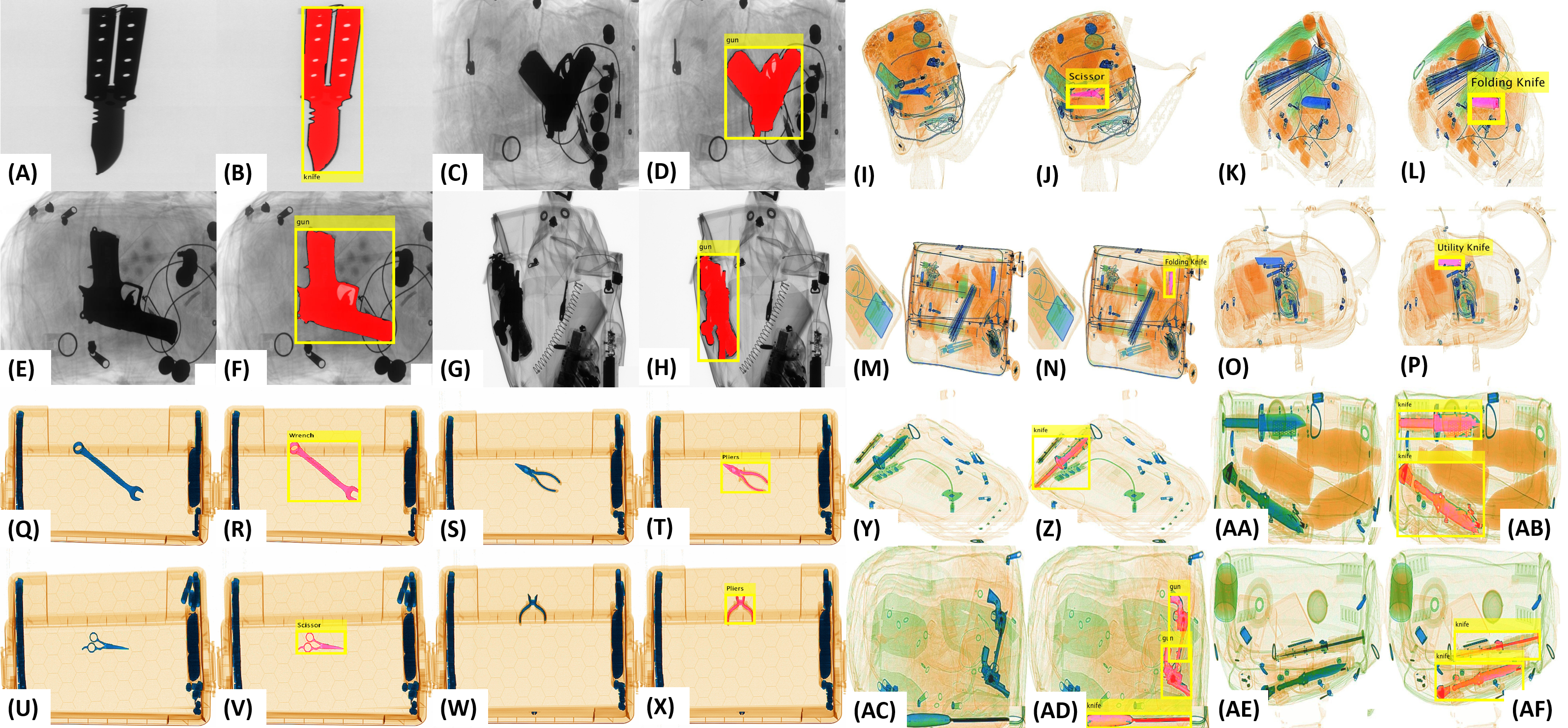}
\end{center}
   \caption{Qualitative evaluation of proposed framework on four public X-ray datasets. (A-H) show scans from the GDXray \RVV{\citep{mery2015gdxray}} dataset, (I-P) show scans from the OPIXray \RVV{\citep{opixray}} dataset, (Q-X) show scans from the COMPASS-XP \RVV{\citep{compass}} dataset, and (Y-AF) show scans from the SIXray \RVV{\citep{miao2019sixray}} dataset.} 
\label{fig:fig5}
\end{figure*}

\noindent Figure \ref{fig:fig5} reports the proposed framework's qualitative evaluations on all four X-ray datasets. We can see here that the proposed framework effectively recognizes the contraband items irrespective of the scanner specifications. However, for cluttered cases, the quality of extracted masks is somewhat limited. For example, see the mask of \textit{gun} in (H). This is because the framework recognizes the anomalous regions from the disparity maps in an unsupervised manner (by clustering their color distributions w.r.t the pixels of the normal baggage content). Therefore, it cannot differentiate the anomalous items' pixel well if they have very high-intensity correlations with the normal pixels. 

\section{Discussion and Conclusion} \label{sec:conclusion}
\noindent This paper presents a novel unsupervised anomaly instance segmentation framework to detect baggage threats from the X-ray scans without any supervision and extensive training procedures. The proposed framework is ideal for screening baggage threats in the real world, easing the security officers' load by avoiding the tedious re-training and ground truth marking process as required in the conventional baggage threat detection frameworks. The proposed framework recognizes the baggage threats as anomalies by exploiting the original and the reconstructed scans' disparities. For cluttered cases, the disparity maps are a bit limited in generating the anomalous items' masks accurately due to their lesser intensity differences with the normal objects. In the future, we plan to improve this aspect by deploying a more adaptive attention mechanism that will highlight only the desired anomalous region within the candidate scan while suppressing the rest of the content. Also, we plan to test the proposed framework to detect 3D-printed and organic baggage threats from the security X-ray scans.   


\section*{Acknowledgements}
This work is supported with a research fund from Khalifa University: Ref: CIRA-2019-047\RV{, and from ADEK Award for Research Excellence: AARE19-156.}

\section*{Authors Contribution Statement}
T.H. devised the idea, wrote the manuscript, and performed the experiments. 
S.A. also contributed to manuscript writing. 
M.B. co-supervised the research and reviewed the experiments. 
S.K. also reviewed the manuscript. 
N.W. supervised the complete research, contributed to manuscript writing, and reviewed the experiments. 

\RV{\section*{Data Availability}
The proposed framework has been thoroughly evaluated on four baggage X-ray datasets, and all of these four datasets are publicly available.
}

\section*{Competing Interests Statement}
All the authors declare that there are no competing interests related to this article.

\end{document}